\documentclass[10pt,journal]{IEEEtran}
\usepackage{amssymb}
\usepackage{graphicx}
\usepackage{picinpar}
\usepackage{booktabs}
\usepackage{float}
\usepackage{amssymb}
\usepackage{url}
\usepackage{cite}
\usepackage{mdwlist}
\usepackage{tikz}
\usepackage{amsfonts}
\usepackage{mathrsfs}
\usepackage{amssymb}
\usepackage{amsmath}
\usepackage{amsthm}
\usepackage{dsfont}
\usepackage{multirow}
\usepackage{color}
\usepackage{graphicx}
\usepackage{subfigure}
\usepackage{subfloat}
\usepackage{epstopdf}
\usepackage{booktabs, tabularx}
\usepackage{array}
\usepackage{slashbox}
\usepackage{algorithmic}
\usepackage[ruled,vlined,linesnumbered]{algorithm2e}

\def\BibTeX{{\rm B\kern-.05em{\sc i\kern-.025em b}\kern-.08em
    T\kern-.1667em\lower.7ex\hbox{E}\kern-.125emX}}
\usepackage{multicol}
\usepackage{microtype}
\usepackage{booktabs} % for professional tables

\usepackage{multirow}
\usepackage{booktabs} 

\usepackage{xcolor}

\usepackage{hyperref}

\ifCLASSINFOpdf

\fi
\begin{document}
\title{Deep Learning Based Unsupervised and Semi-supervised Classification for Keratoconus}

\author{Nicole Hallett, Kai Yi, Josef Dick, Christopher Hodge, Gerard Sutton, Yu Guang Wang$^{*}$, Jingjing You$^{*}$

\thanks{N. Hallett, C. Hodge, G. Sutton and J. You are with the Sydney Eye Hospital, The University of Sydney, Sydney, Australia (emails: nhal6992@uni.sydney.edu.au;
christopher.hodge@vei.com.au;
gerard.sutton@vei.com.au;
jing.you@sydney.edu.au).}
\thanks{K. Yi, J. Dick and Y. G. Wang are with the School of Mathematics and Statistics, The University of New South Wales, Sydney, Australia (e-mails:
kai.yi@student.unsw.edu.au;
josef.dick@unsw.edu.au;
yuguang.wang@unsw.edu.au).}
\thanks{*~Corresponding author}}
\markboth{}  %
{\MakeLowercase{\textit{et al.}}: }
% The only time the second header will
\maketitle

\begin{abstract}
The transparent cornea is the window of the eye, facilitating the entry of light rays and controlling focusing the movement of the light within the eye. The cornea is critical, contributing to 75\% of the refractive power of the eye. Keratoconus is a progressive and multifactorial corneal degenerative disease affecting 1 in 2000 individuals worldwide. Currently, there is no cure for keratoconus other than corneal transplantation for advanced stage keratoconus or corneal cross-linking, which can only halt KC progression. The ability to accurately identify subtle KC or KC progression is of vital clinical significance. To date, there has been little consensus on a useful model to classify KC patients, which therefore inhibits the ability to predict disease progression accurately.

In this paper, we utilised machine learning to analyse data from 124 KC patients, including topographical and clinical variables. Both supervised multilayer perceptron and unsupervised variational autoencoder models were used to classify KC patients with reference to the existing Amsler-Krumeich (A-K) classification system. Both methods result in high accuracy, with the unsupervised method showing better performance. The result showed that the unsupervised method with a selection of 29 variables could be a powerful tool to provide an automatic classification tool for clinicians. These outcomes provide a platform for additional analysis for the progression and treatment of keratoconus.
\end{abstract}


\IEEEpeerreviewmaketitle

\section{Introduction}
Artificial intelligence (AI) is an emerging field in ophthalmic science and medicine. While many researchers have utilised deep and machine learning to monitor and analyse ocular diseases such as diabetic retinopathy \cite{1-date2019applications}, macular degeneration \cite{2-motozawa2019optical} and glaucoma \cite{3-ahn2018deep}, to date little AI research has been done within the study of human corneal disease.

Keratoconus (KC) is a bilateral, asymmetric, progressive corneal disease, affecting some 1 in 2,000 patients worldwide (Pearson et al. 2000). It is characterised by central and para-central corneal thinning, leading to induced myopia and irregular astigmatism causing deterioration of the patient's best-corrected visual acuity \cite{4-serdarogullari2013prevalence}. Currently, there is little consensus on the etiology of KC however, both genetic and environmental risk factors are considered to play a role \cite{5-tur2017review}. Research suggests a complex matrix of risk factors including gender, age, atopy, sun exposure, geography, allergies, eye rubbing, contact lens wear, dominant sleeping side, body mass index, amongst others. Outside eye rubbing, however, the presence of other factors have provided contradicting findings within the literature \cite{6-davidson2014pathogenesis,mcmonnies2015inflammation,gordon2015risk}.

The main diagnostic tool currently utilised in KC is corneal topography and tomography, which describes the surface curvature of the cornea. Various methods to obtain topographical values exist, including Placido ring and Scheimpflug camera measurement. The methodology impacts the available metrics. Supplementary diagnostic technology may be used including Optical coherence tomography (OCT), which utilises low coherence interferometry to produce a two or three-dimensional image. This may provide additional quantitative and qualitative information on aspects such as corneal thickness and both anterior chamber angle and depth information \cite{8-li2012anterior}. Although topography and tomography represent a key diagnostic indication, it is noted that a single diagnostic factor is not sufficiently accurate to confirm an early diagnosis or indeed, disease progression \cite{7-martinez2017new}.

Corneal transplantation replaces the diseased cornea with donor tissue and may restore corneal regularity and best-corrected visual acuity. However, the surgery and postoperative process is not without significant risks and considered a final treatment option only. The most effective intervention currently available to halt the progression of KC is corneal cross-linking (CXL). Cross-linking is indicated in the presence of disease progression, identified by a combination of increasing corneal curvature, irregularity and refractive changes.  The decision to proceed to CXL represents a clinical challenge as patients will progress at different rates.

Similarly, not all patients will progress to require corneal transplantation for visual rehabilitation. Although routinely valid, surgical complications have been identified and CXL does not appear to work in between 5-20\% of procedures \cite{ozer2019long}. Projecting both the likelihood and rate of progression against the risks of surgical intervention is, therefore, key to optimising the timing of CXL intervention and the potential refractive and visual outcomes. Consequently, an improved model of progression prediction may prove invaluable.

To achieve an accurate model of prediction, it remains essential to classify patients and the stages of the disease process. Currently, no universal classification scheme exists for keratoconus \cite{15-li2009keratoconus}. The purpose of this pilot study, which is the first step in developing an accurate model of disease progression prediction, is to accurately classify KC patients through the use of machine learning.

This research aims to examine machine learning models through the use of patient data to develop a model that can accurately classify different stages of KC, with two key objectives:
\begin{itemize}
    \item To use supervised and unsupervised models to classify different stages of KC;
    \item To compare the accuracy of the supervised and unsupervised models in the classification of KC.
\end{itemize}

The paper is organised as follows. In Section~\ref{sec:litereview}, we review the related work on Keratoconus diagnosis by machine learning methods. Section~\ref{sec:dataset} describes a keratoconus patient data set collected in a private ophthalmic clinic (Vision Eye Institute (VEI) Chatswood), which is used in this study. In Section~\ref{sec:vae}, we develop a variational autoencoder (VAE) with Gaussian mixture classifier to cluster the corneal data into four A-K classes to reflect the severity of KC in our cohort. The VAE with application to our corneal data set demonstrates excellent performance with clustering accuracy. In Section~\ref{sec:mlp}, we develop a multilayer perceptron model to predict the A-K classification label from known labels. The MLP has state of the art performance on the real corneal data. In Section~\ref{sec:conclusion_discussion}, we outline the results of the study, as well as the next steps for this research.

\section{Literature Review}\label{sec:litereview}
Deep learning has significantly improved object recognition and medical image analysis \cite{10-alio2017keratoconus}. In recent years there has been an increase in research with deep learning and ocular disease.

The aim of machine learning in KC research to date has predominantly been to identify early or subclinical forms of the disease. A recent review by Lin et al. \cite{11-lin2019review} details 17 publications concerning the detection of KC, 15 of the 17 reviewed papers utilised topographic maps as the primary input, with 65\% of the papers seeking to classify KC patients, and 30\% seeking to distinguish KC or sub-KC. While some of the reviewed papers utilised multiple machine learning models, 53\% utilised neural networks, 23\% utilised discriminant analysis. Differentiation across accuracy, sensitivity and specificity was apparent in the reviewed papers with ranges reported between 65.2\%-100\%, 63\%-100\% and 82\%-100\% respectively.

Of most relevance to this study, several cohorts utilised the Pentacam corneal topography and tomography unit, thereby allowing objective comparisons. See \cite{12-hwang2018distinguishing,13-kovacs2016accuracy,14-yousefi2018keratoconus}. Further, Yousefi et al. \cite{14-yousefi2018keratoconus} used an unsupervised algorithm to cluster KC associated variables with the greatest accuracy effectively.

Kovacs et al. \cite{13-kovacs2016accuracy} undertook a study with a Pentacam HR Scheimpflug device. This retrospective case-control study utilised a multilayer perceptron neural network to assess the corneal symmetry in 60 eyes from 30 patients who had unilateral keratoconus. This research classified patients based on videokeratography and clinical signs through the framework of KISA, the three components of which are central keratometry (K), the inferior-superior (I-S) value and keratoconus percentage index \cite{15-li2009keratoconus}. The accuracy of machine learning classification in this study was based on the Pentacam progress index (PPI). However, while this functionality is readily available on Pentacam devices, it has not widely been used to classify KC within the existing literature.

Kovacs et al. used classifiers trained on variables through the creation of bilateral data for each parameter. The best neural network architecture was determined from a feedforward network based on highest accuracy, through the training and test sets of 70\% and 30\% respectively. Results indicated the highest accuracy was to identify subtle corneal changes in unilateral KC patient's control eyes which have both sensitivity and specificity at 90\%. However, while the study demonstrated high accuracy between normal, subnormal and KC groups, it did not discriminate between severity levels of KC within the KC group. Further, as the control group were represented by the opposite and previously non-diagnosed eye, the practical benefit of these findings remains unclear. KC is considered bilateral disease albeit often highly asymmetrical in the presentation. Confirmation of a diagnosis of KC in the less affected eye is therefore to be expected. 

Hwang et al. \cite{12-hwang2018distinguishing} utilised multivariate logistic regression analysis and hierarchical algorithm to determine the optimal objective, machine derived variables and combinations, utilising an approach of combining metrics from two devices Pentacam and Spectral Domain OCT. 
This retrospective case-control study analysed asymmetric clinically normal fellow eyes from 30 KC patients and 60 clinically normal eyes from 60 bilaterally normal control patients.
While the authors did not specify training and testing set sizes, in establishing sensitivity, specificity and area under curve (AUC), it demonstrated the ability to reduce the 24 variables to 13, utilising the hierarchical clustering method, highlighting the benefits of utilising multiple devices and inputs.
Results were highest for the clustered combination from both the Pentacam and OCT device, with the combined 13 variables as listed in Table~\ref{tab:pentacam_oct} below resulting in 100\% sensitivity and 100\% specificity distinguishing KC from control corneas.

\begin{table}[ht]
\begin{minipage}{\columnwidth}
\caption{Pentacam and OCT combined variables from \cite{12-hwang2018distinguishing}}
\label{tab:pentacam_oct}
\begin{center}
\begin{tabular}{l|l}
\toprule
Pentacam and & \multirow{2}{*}{Resultant 13 variables}\\ 
OCT combined &\\
\midrule
  &  Temporal Outer\\
 & Pachymetry Minimum\\
  &  Temporal Inner\\
  & Index Vertical Asymmetry\\
 & Central Thickness\\ 
 & Epithelium Standard deviation\\ 
 & Minimum - Medium SD-OCT\\ 
 & Index Surface Variance\\ 
 & Inferior Temporal Inner\\
 & Epithelium Minimum-maximum\\
 & Minimum SD-OCT\\ 
 & Superior Outer \\
 & Superior Inner \\
\bottomrule
\end{tabular}
\end{center}
\end{minipage}
\end{table}

Yousefi et al. \cite{14-yousefi2018keratoconus} utilised unsupervised machine learning to predict KC severity. The study utilised Swept-Source OCT images (CASIA), to analyse 3,000 eyes representing a significant data set. Unlike most studies the Ectasia Severity Index (ESI) was used to determine participant suitability. The ESI, which analyses corneal changes and degeneration is an instrument guided screening index, is not routinely used in clinics minimising the practical impact of the research. 

This study utilised Swept-Source OCT images (Casia) of 12,242 eyes from multiple centres in Japan, from which 3,156 eyes with valid ESI were selected. Of interest, the paper only outlines patient selection based on ESI and does not include details on patient history. The algorithm is comprised of 3 key steps, principal component analysis (PCA) to reduce input data from 420 to 8 significant components linearly; manifold learning to reduce non-linearly parameters to eigen-parameters; density-based clustering to identify keratoconic eyes. 

This study utilised t-SNE to consider clustering of variables within the analysis, resulting in four different clusters of patients,  reflecting the classification of patients. The study achieved a specificity and sensitivity of identifying healthy eyes from KC eyes of 94.1\% and 97.7\% respectively. The results of this study distinguished KC from normal eyes, rather than establishing a well-defined model of stages within KC.

While emerging, most of the machine learning studies with KC have been undertaken, focusing primarily on the classification of pachymetric images. To date, machine learning has not been utilised to establish an association between the multifactorial variables of KC.

In the present study, Pentacam image data combined with risk factors such as age, gender, eye rubbing and others were included. Both supervised, and unsupervised machine learning was used and compared to a widely used classification system: Amsler-Krumeich (A-K) classification system. The primary goal is to accurately classify KC patients, to develop a more accurate model for early disease progression prediction, rather than identifying or classifying KC from subclinical and normal samples.

\section{Patient corneal data set}\label{sec:dataset}
This is a retrospective single centre study approved by the University of Sydney Human Research Ethics Committee (HREC 2013/1041). Patient data from 124 KC patients was collected from Vision Eye Institute Chatswood between 2014- 2017. The medical records of each patient were reviewed and analysed, of which 79\% (63.7\%) were male. The information in Table~\ref{tab:variables} was extracted to be utilised as variables of consideration in both supervised and unsupervised models.

\begin{table}[ht]
\begin{minipage}{\columnwidth}
\caption{Variables utilised in study}
\label{tab:variables}
\begin{center}
\begin{tabular}{l|l}
\toprule
Variable &	Source\\\midrule
Gender &	Patient questionnaire\\
Age	& Patient questionnaire\\
Nationality &	Patient questionnaire\\
Diabetes &	Medical record\\
Atopy	& Medical record\\
Allergy	& Medical record\\
Hypertension & Medical record\\
Other disease presence & Medical record \\
Length of time since diagnosis	& Patient questionnaire\\
Known eye history &	Patient questionnaire\\
Family history &	Patient questionnaire\\
Eye rubbing 	& Patient questionnaire\\
Primary optical aid	& Patient questionnaire\\
Uncorrected distance visual acuity & VEI clinician assessment\\
Corrected distance visual acuity
	& VEI clinician assessment\\
Presence of hydrops	& VEI clinician assessment\\
Corneal scarring & VEI clinician assessment\\
Vogt's Striae & VEI clinician assessment\\
Fleischer's ring & VEI clinician assessment\\
Refractive sphere &	Pentacam\\
Refractive cylinder &	Pentacam\\
Refractive axis	& Pentacam\\
Flat keratometry &	Pentacam\\
Steep keratometry &	Pentacam\\
Thinnest pachymetry & Pentacam\\
Location X axis &	Pentacam\\
Location y axis &	Pentacam\\
Central pachymetry & Pentacam\\
Amsler-Krumeich (AK) classification & VEI clinician \\
\bottomrule
\end{tabular}
\end{center}
\end{minipage}
\end{table}

All patients were classified, through clinician experience, based on the Amsler-Krumeich (A-K) classification, which is comprised of mean-K readings on the anterior curvature sagittal map, thickness at the thinnest location and the refractive error of the patient, and biomicroscopy. Table~\ref{tab:AK} \cite{kamiya2014evaluation} shows how A-K classification groups patients. 

\begin{table}[ht]
\begin{minipage}{\columnwidth}
\caption{A-K classification by \cite{kamiya2014evaluation}}
\label{tab:AK}
\begin{center}
\begin{tabular}{l|l}
\toprule
Grad & Characteristic\\
\midrule
 \multirow{3}{*}{1} &  Eccentric steeping\\
& Myopia and astigmatism $<5.00$D\\
& Mean central K readings $<48.00$D\\
\midrule
\multirow{4}{*}{2} & Myopia and astigmatism 5.00-8.00D\\
& Mean central K readings $<53.00$D\\
& Absence of scarring\\
& Minimum corneal thickness $>400\mu$m\\
\midrule
\multirow{4}{*}{3} & Myopia and astigmatism 8.00-10.00D\\
& Mean central K readings $>53.00$D\\
& Absence of scarring\\
& Minimum corneal thickness 300-400$\mu$m
\\
\midrule
\multirow{4}{*}{4} & Refraction not measurable\\ 
& Mean central K readings $>55.00$D\\
& Central corneal scarring\\
& Minimum corneal thickness $200\mu$m
\\
\bottomrule
\end{tabular}
\end{center}
\end{minipage}
\end{table}

%Of significance is that our study groups patients based on clinician experience for the classification. 
%
%Data was analysed through 2 different machine learning models, one supervised and one unsupervised. We used a supervised multi-layer perceptron (MLP) model with three hidden layers and network architecture of 29-128-256-4. For the unsupervised model we utilised a Bayesian deep neural network with two hidden layers and an architecture of encoder and decoder of 29-128-256-2-1 and 1-256-129-29 respectively. Both models were run with an epoch of 100.

\section{Clustering for Corneal Data by Variational Autoencoder}\label{sec:vae}
% \subsection{t-SNE}
Variational autoencoder (VAE) \cite{16-kingma2013autoencoding} is a Bayesian deep neural network, which consists of an encoder and a decoder and a latent variable layer. The encoder and decoder which are deep neural networks are used to extract features from the input data and to generate the same type of output data from the latent features respectively. The encoder is a deep net built to learn features of the input data which are then passed to the latent variable layer. In the decoder, the latent features are used to generate the output data with the same format of the original data. Between the encoder and decoder, the latent variable layer uses Gaussian random sampling to generate latent features. The clustering is then obtained by features from the latent variable layer between the encoder and decoder.

\begin{figure}[htbp]
\begin{minipage}{\columnwidth}
\centering
    \includegraphics[width=\columnwidth,height = 3cm]{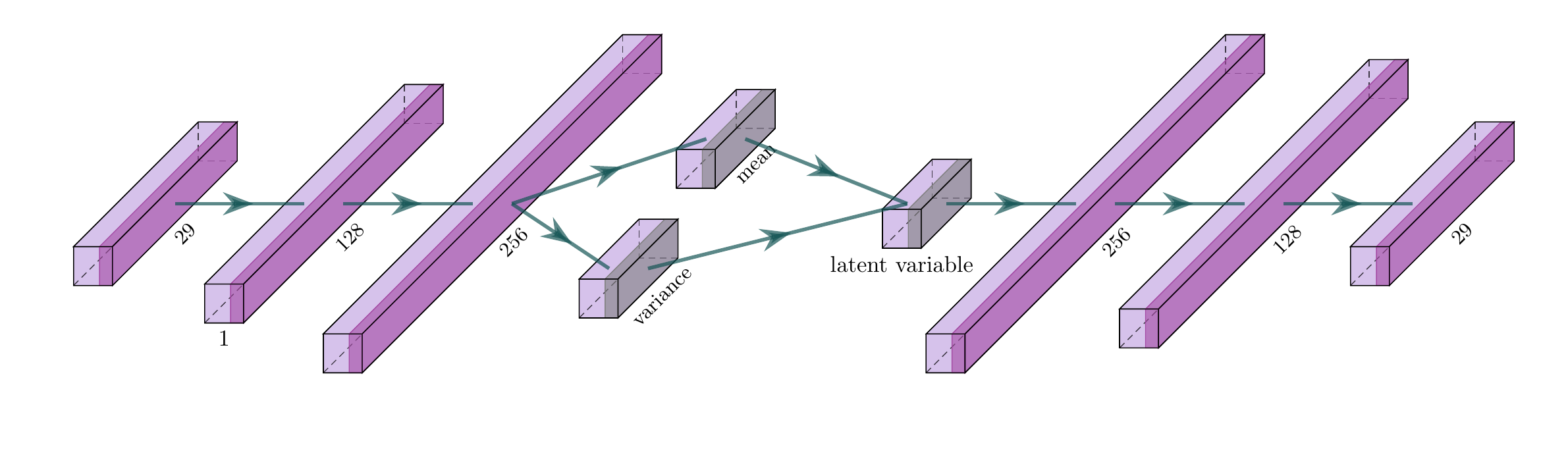}
\end{minipage}\vspace{-5mm}
\begin{minipage}{\columnwidth}
\caption{VAE architecture: encoder, decoder and latent variable layer; the encoder and decoder are MLP with 2 hidden layers.}\label{fig:vae}
\end{minipage}
\end{figure}

In this work, we take the encoder and decoder as multilayer perceptron (MLP) models, which is a fully connected deep neural network, see Section~\ref{sec:mlp}. The input data was trained by the VAE of the above network architecture (see Figure~\ref{fig:vae}). The encoder uses a deep net to compress each high-dimensional input sample into a two-dimensional real vector. By this, the encoder extracts the features of the input and is then clustered by a Gaussian mixture model \cite{17-dilokthanakul2016deep} to a given number of classes.

Figure~\ref{fig:vae} shows the VAE model we use in the experiments. The encoder and decoder are MLP with 2-hidden layers. The network architecture of encoder and decoder are 29-128-256-2 and 2-256-128-29, where the $29$ is the number of the variables in the corneal data, $256$ and $128$ are the numbers of hidden neurons of the deep nets. These are hyper-parameters which have been tuned to optimize the performance of the deep network. When the VAE model has been trained, a Gaussian mixture model is applied to cluster the compressed 2D vectors.

\subsection{Bayesian inference in VAE}
The VAE is interpreted as a Bayesian inference model. The $N$ input data of encoder $X=\{x^{(i)}\}_{i=1}^N$ and $N$ outputs of decoder $Y=\{y^{(i)}\}_{i=1}^N$ and latent feature $z$ are regarded as random variables which are in the probability spaces $\mathcal{X},\mathcal{Y}$ and $\mathcal{Z}$.
The encoder is characterised by the conditional probability density (CPD) $q_{\phi}(z|x)$ satisfying
\begin{equation*}
    p(z) = \int_{\mathcal{X}}q_{\phi}(z|x) p_{X}(x)\mathrm{d}x,
\end{equation*}
where $p_X$ is the (unknown) probability density function of the input data. The $\phi$ signifies the parameters of the encoder neural network.

The decoder is depicted by the CPD $p_{\theta}(y|z)$, which links the latent variable $z$ by
\begin{equation*}
    p_{\theta}(y) = \int_{\mathcal{Z}}p_{\theta}(y|z)p(z) \mathrm{d}z \quad \forall y\in Y,
\end{equation*}
where $p$ is the probability density function of the latent variable and $\theta$ signifies the parameters of the decoder neural network.

The marginal log-likelihood of output data set $Y$ is given by $\log p_{\theta} (Y) := \sum_{i=1}^N \log p_{\theta}(y^{(i)})$. For  sample $i=1,\ldots, N$, the log-likelihood $\log p_{\theta}(y^{(i)})$ has a variational lower bound \cite{mackay2003information}
%\begin{equation}
\begin{align*}%\label{eq:ELBO}
    \log p_{\theta}\left(y^{(i)}\right) 
    &\geq \mathbf{E}_{z}\left[\log p_{\theta}\left(y^{(i)} | z\right)\right] 
    \\ \nonumber
    &\qquad - D_{\rm KL}\left(q_{\phi}\left(\cdot | x^{(i)}\right) \| p( \cdot )\right),
\end{align*}
where the $D_{\mathrm{KL}}$ term represents the Kullback-Leibler divergence \cite{kullback1951information} of the posterior distribution of the output of the encoder and the prior distribution of the latent variable, and the $-\mathbf{E}_{z}$ term is the reconstruction error of the decoder output. From this, we can define the loss function of the VAE as
\begin{align*}
    \mbox{LOSS} &= \sum_{i=1}^N \biggl\{D_{\mathrm{KL}}\left(q_{\phi}(\cdot |x^{(i)})\| p_{\theta}(\cdot )\right) \\
    &\quad\qquad\qquad - \mathbf{E}_{z}\left[\log p_{\theta}(y^{(i)}|z)\right]\biggr\}.
\end{align*}

In training the neural network, the VAE uses back-propagation to minimize the loss function.
The $q_{\phi}(\cdot |x)$ and $p_{\theta}( \cdot )$ are set as normally distributed density functions. The mean and variance of $q_{\phi}(\cdot |x)$ are parameters by a deep neural network, and the $p_{\theta}(\cdot )$ is the standard normal distribution density by which the latent variable $z$ is sampled.

\subsection{Experimental results and clinical interpretation} 
\begin{figure}[htbp]
    \centering
    \includegraphics[width = \columnwidth]{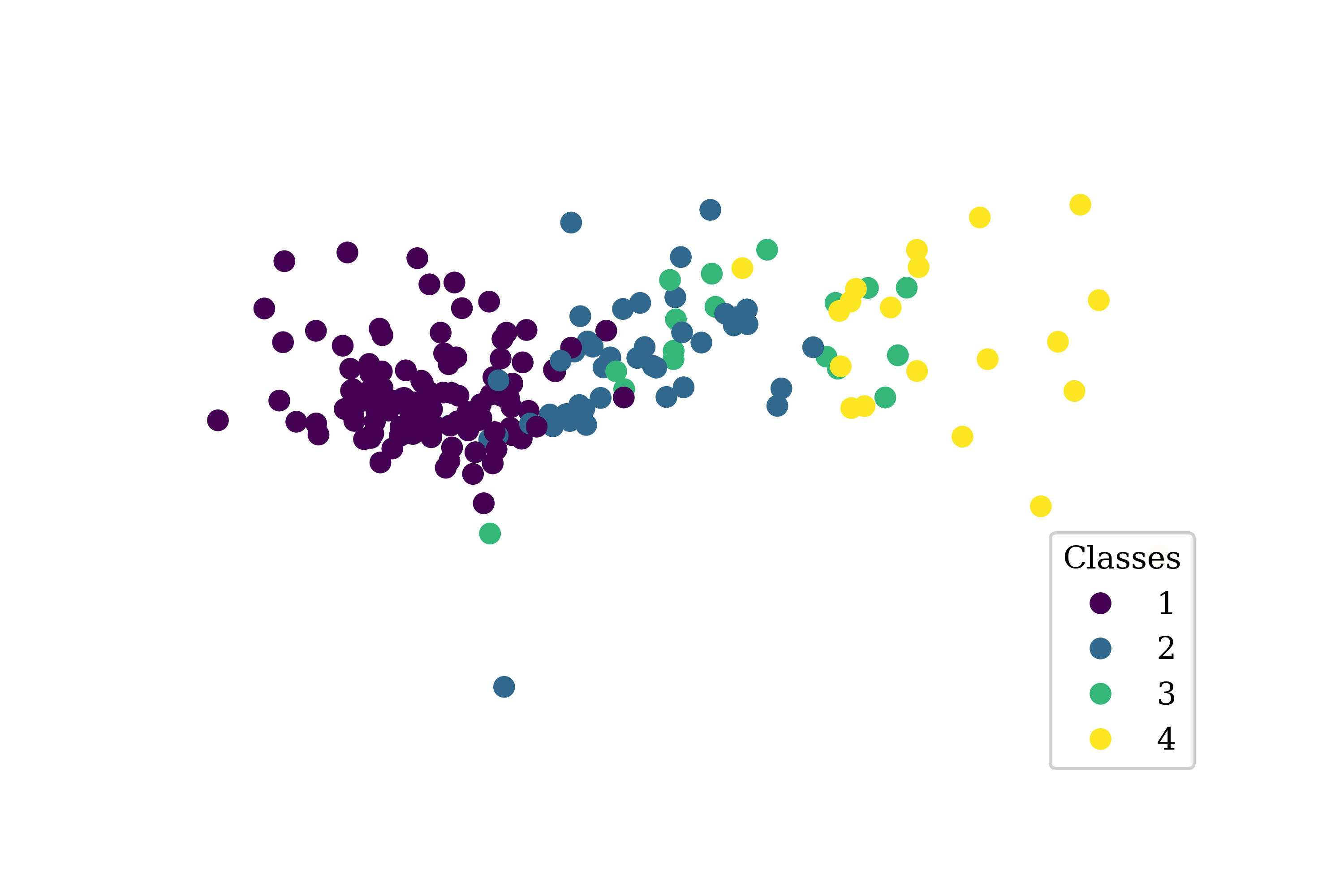}\vspace{-6mm}
    \caption{The corneal data is compressed into two dimensional vector by VAE. They are the input of the Gaussian mixture model for clustering. The 2D vectors are the extracted features of the input data which can then be used for clustering. Here, the class label for each sample is from the ground truth.}\label{fig:vae_result_1}
\end{figure}

\begin{figure}[htbp]
    \centering
    \includegraphics[width = \columnwidth]{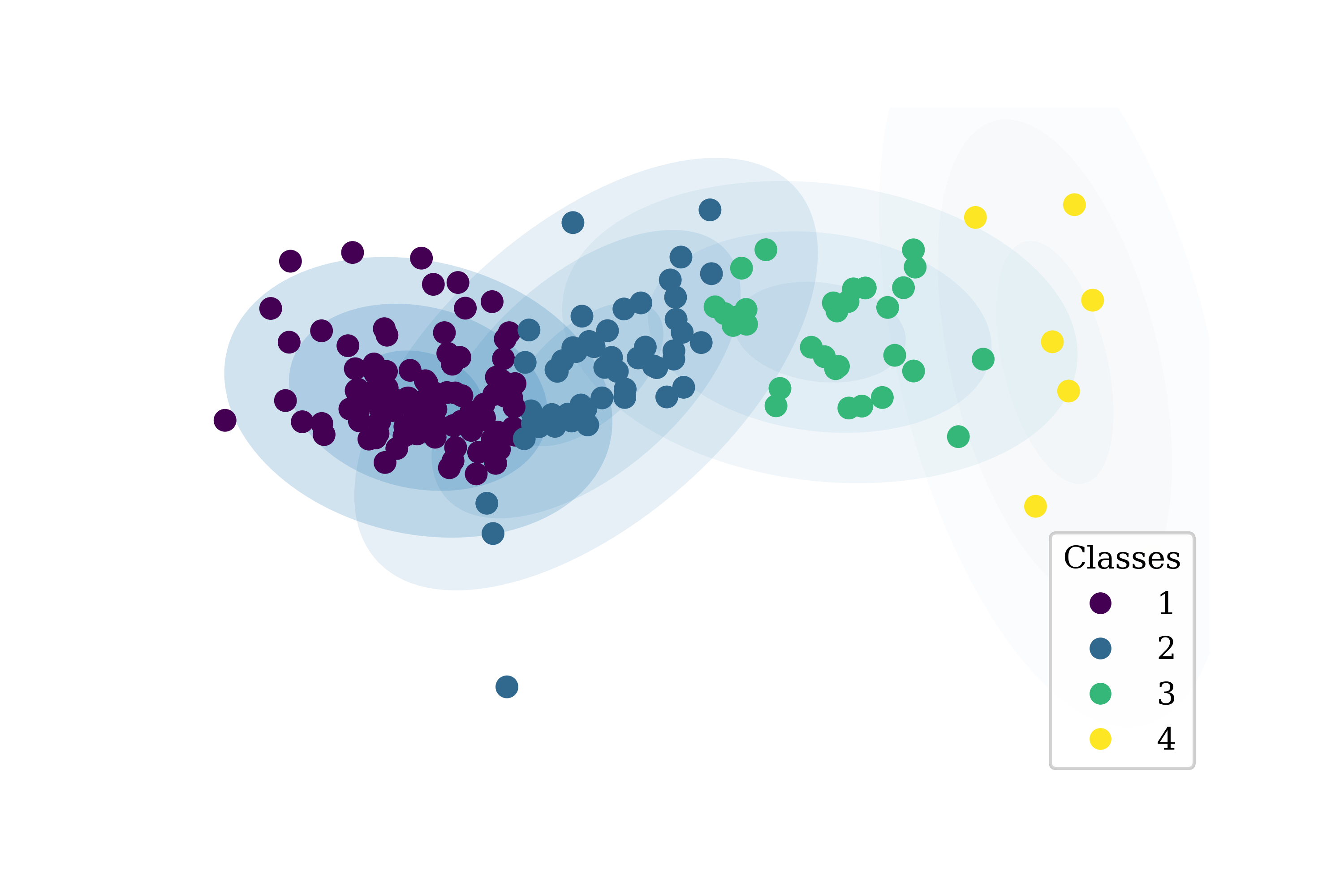}\vspace{-6mm}
    \caption{Corneal data is clustered into 4 classes by Gaussian mixture model using the latent 2D vectors from the VAE model. The clustering accuracy compared to the ground truth is 80.3\%. The shallow blue ellipses show the confidence region of the centre of each cluster. The difference between class 1 and the classes 3 and 4 are apparent; the 1st and the 2nd classes are close to each other.}\label{fig:vae_result_2}
\end{figure}

As mentioned, we use VAE to reduce the dimension of input data from the 29 initial variables to 2. We then utilise Gaussian mixture modelling for clustering to classify the compressed sample data into one of the 4 A-K classes. Figure~\ref{fig:vae_result_1} shows the plot of latent 2D feature vectors of the corneal input data. The class label is determined by the AK classification label of the ground truth. We can observe that there is already an embryonic form of clustering, but the classification is not clear as different clusters have many overlaps.  

We thus send the latent 2D feature vectors to the Gaussian mixture model, which then clusters the samples into 4 classes.
As depicted in Figure~\ref{fig:vae_result_2}, the classes of 1,2,3 and 4 relate to A-K classifications of 1-4 as outlined in the introduction. Patients in classes 1 and 4 are independent, as the groups have different mean and variance in $R^2$. The results also indicate distinct features between classes 1 and 3. Our results infer that classes 1 and 2 are the most similar, as evidenced with overlapping on the cluster plot, which is brought about by miss-clustering of the data for patients classified within groups 1 and 2. The accuracy of the VAE in Figure~\ref{fig:vae_result_2} compared with the ground truth is as high as 80.3\%. In 20 repetitions of the test, comparing these results to the ground truth of A-K classification through clinical diagnosis represents a significant outcome with accuracy at 76.9\% with Std Dev. of 3\%, and the highest accuracy level of 82.4\%. The Std Dev. 3\%, which is small, reflects the uncertainty in the sampling of the latent variable and Gaussian mixture clustering of the VAE.

\begin{figure}[htbp]
    \centering
    \includegraphics[width = \columnwidth]{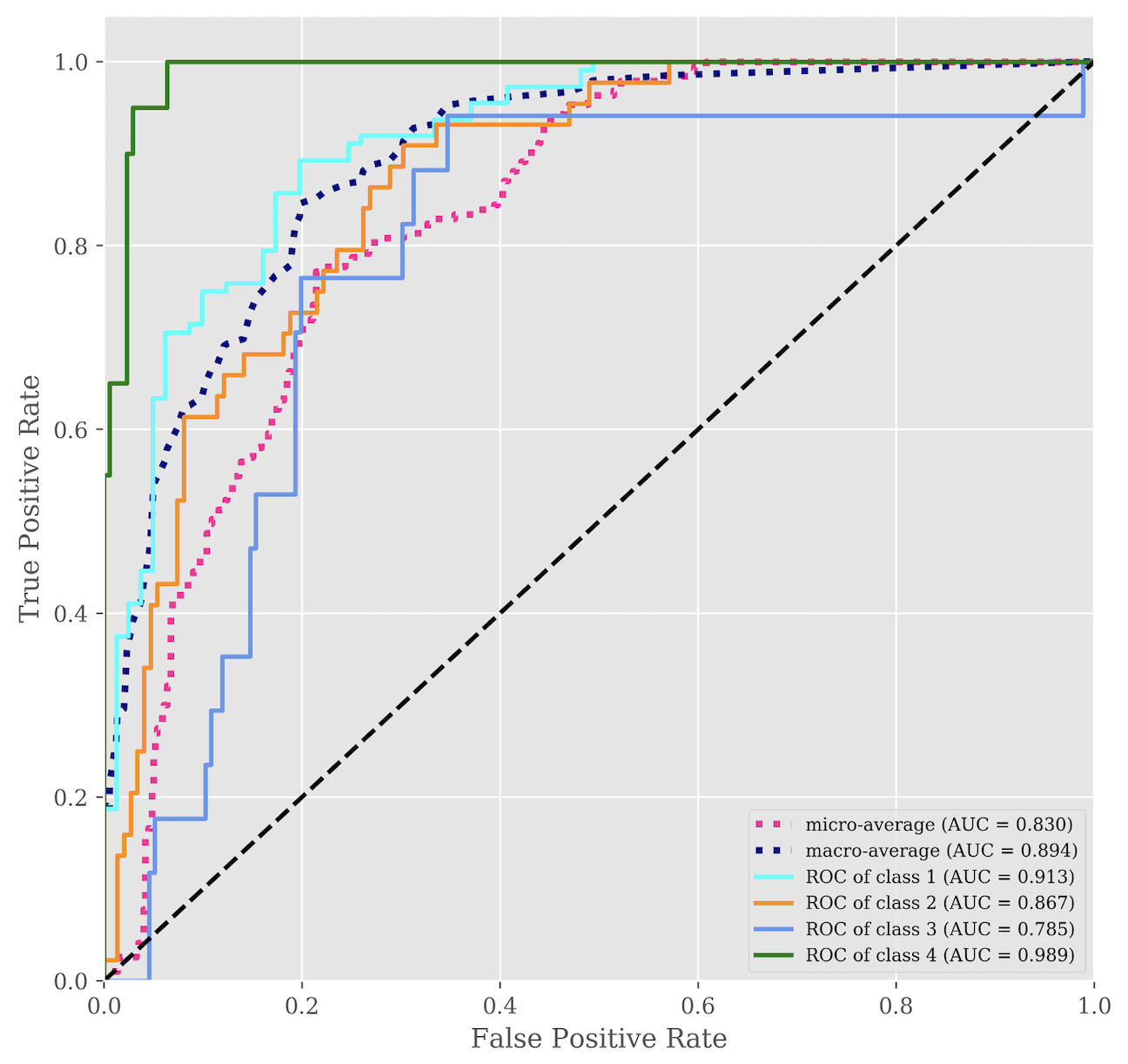}
    \caption{ROC curves for VAE classifier. The VAE classifier shows excellent performance on each of four classes, with AUC between 0.79 to 0.99.}\label{fig:vae_roc}
\end{figure}

Figure~\ref{fig:vae_roc} shows the ROC curves and AUC for the VAE classifier. The ROC curves for all classes are close to the upper left corner, and the AUC for classes 1,2,3 and 4 achieves high values at 0.91, 0.87, 0.79 and 0.99. It illustrates that the VAE classifier has excellent performance for the corneal data clustering.
Meanwhile, the VAE has helped to classify patients into 4 classes that correspond to the A-K classification. As this is an unsupervised model, it will facilitate clinicians being able to accurately group patients with an A-K classification based on the input 29 variables of real measurement. 

\section{Deep Neural Networks for Corneal Diagnosis}\label{sec:mlp}
We use the multi-layer perceptron (MLP) model for corneal data classification. The MLP model is a fully connected deep neural network with multiple layers for a semi-supervised learning task, see \cite{18-lecun2015deep,19-Goodfellow-et-al-2016}.
\begin{figure}[htbp]
\begin{minipage}{\columnwidth}
\centering
    \includegraphics[width=0.8\columnwidth,height = 2.8cm]{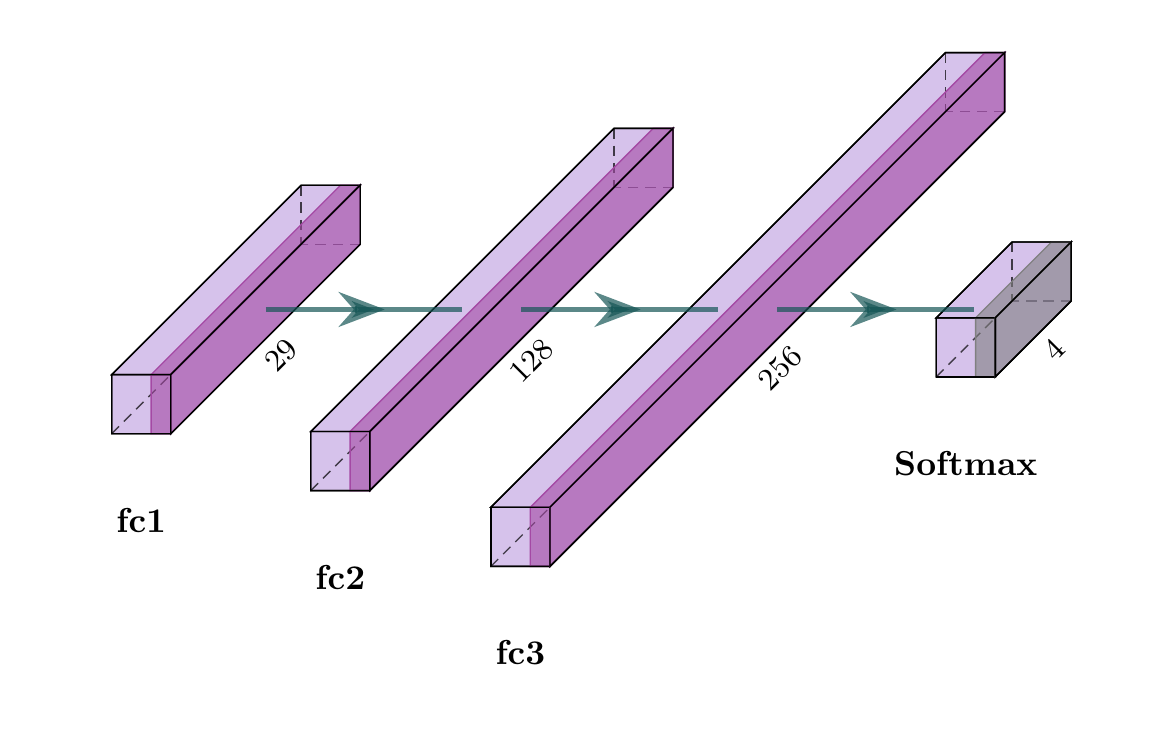}
\end{minipage}
\begin{minipage}{\columnwidth}
\caption{Network architecture of MLP with three hidden layers. It has 128, 256 and 4 neurons in the second, third and fourth hidden layers.}\label{fig:mlp}
\end{minipage}
\end{figure}

\subsection{Multilayer perceptron}
An MLP of depth $K$ is a multivariate function in the form of
\begin{equation*}
	x^{(j+1)}_i	= g_{ij}(x^{(j)}\cdot w_{ij} + b_{ij}),
\end{equation*}
for $j=1,2,\dots,K$ and $j=1,2,\dots,N_j$,
where the $x^{(j)}=(x^{(j)}_1,\dots,x^{(j)}_{N_j})$ is the input of $j$th layer with $N_j$ neurons, and $g_{ij}$ is the activation function on $\mathbb{R}$ which connects the input and the $i$th neuron in the $j$th layer, and $w_{ij}$ and $b_{ij}$ are the trainable weights and biases. For classification task, the last MLP is trained by back-propagation using a stochastic gradient descent optimisation strategy.

\begin{figure}[htbp]
\centering
\begin{minipage}{0.9\columnwidth}
\centering
    \includegraphics[width=\columnwidth]{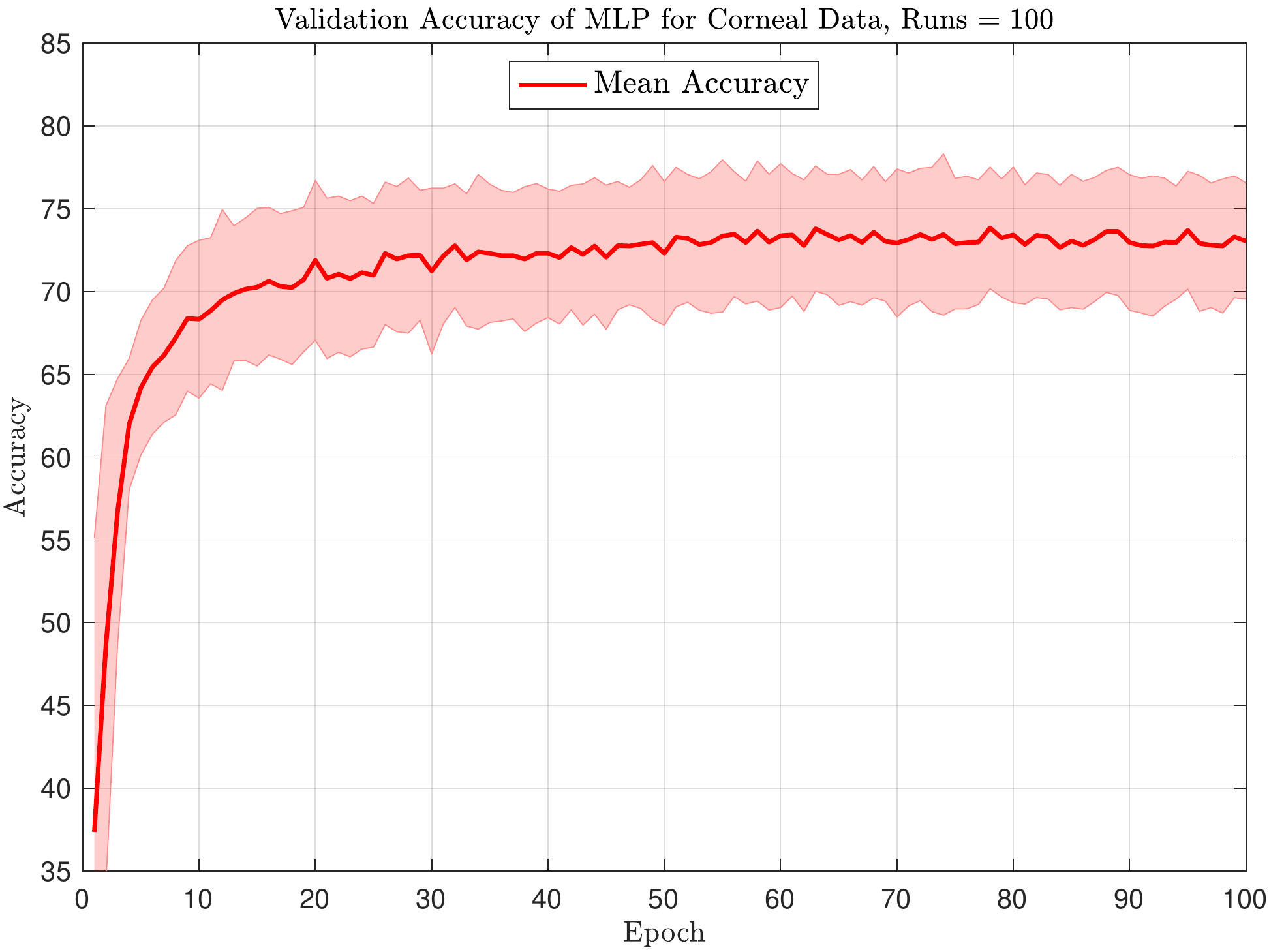}
\end{minipage}
\begin{minipage}{0.9\columnwidth}
\caption{Validation accuracy of MLP for corneal data; the red curve shows the mean validation accuracy with epoch up to 100 in training; the shallow red region shows the variance of validation accuracy with epoch.}\label{fig:val_acc}
\end{minipage}
\end{figure}

\begin{figure}[htbp]
\centering
\begin{minipage}{0.9\columnwidth}
\centering
    \includegraphics[width=\columnwidth]{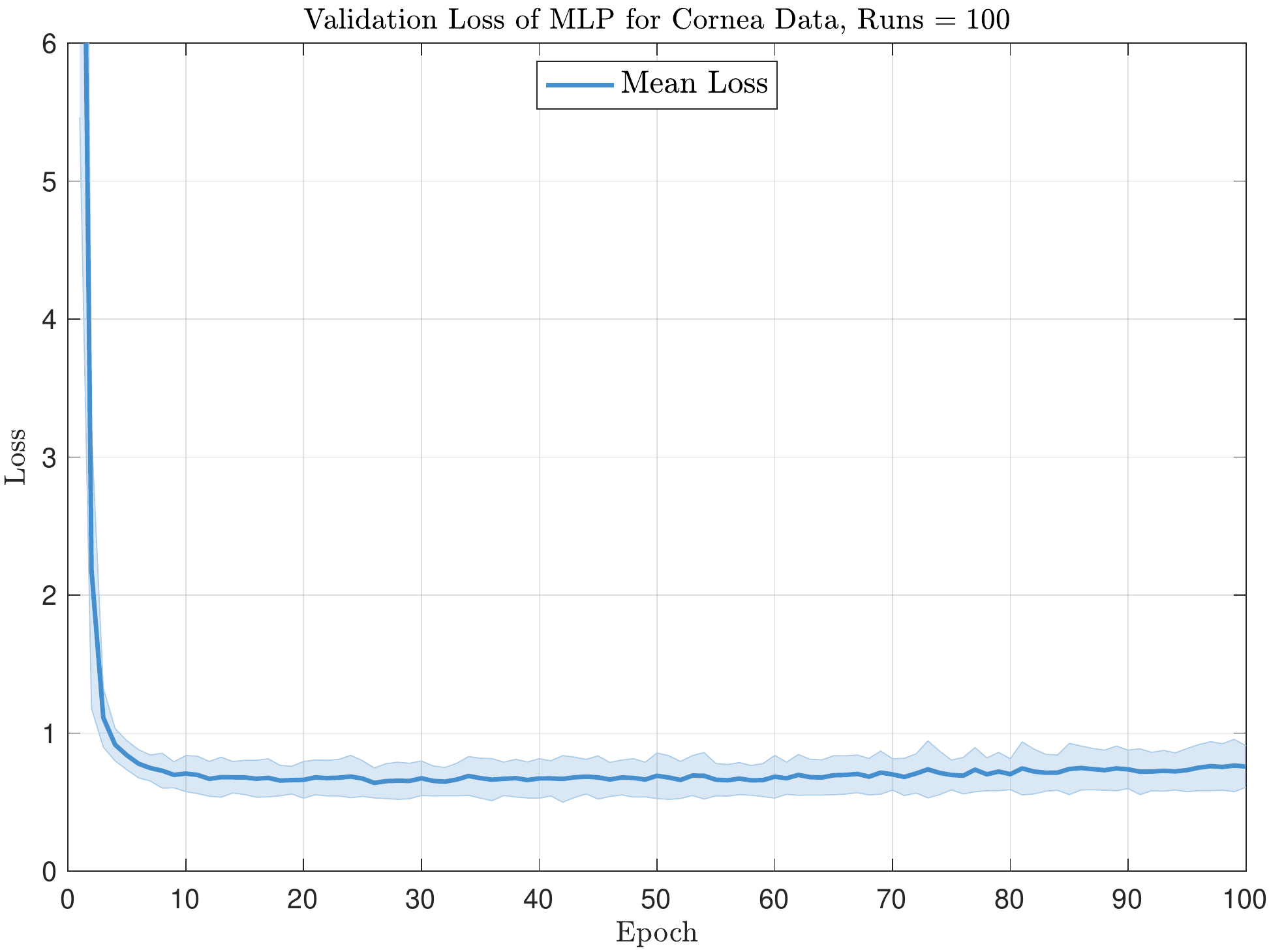}
\end{minipage}
\begin{minipage}{0.9\columnwidth}
\caption{Validation loss of MLP for corneal data; the blue curve shows the mean validation loss with epoch up to 100 in training; the shallow blue region shows the variance of validation loss with epoch.}\label{fig:val_loss}
\end{minipage}
\end{figure}

\begin{figure}[htbp]
    \centering
    \includegraphics[width = \columnwidth]{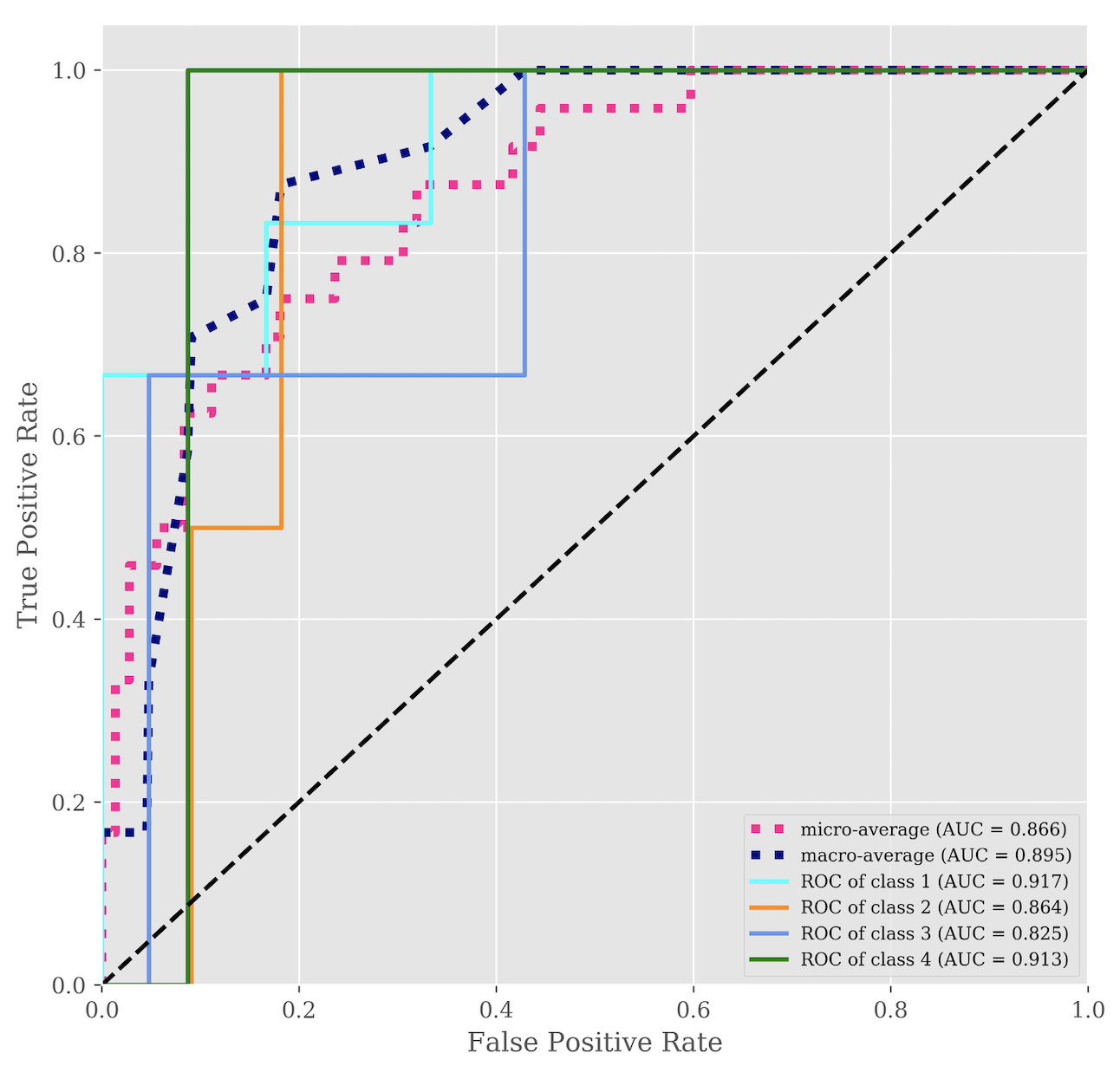}
    \caption{ROC curves for MLP classifier. The MLP classifier shows excellent performance for each of four classes, with AUC between 0.83 and 0.92.}\label{fig:mlp_roc}
\end{figure}

Figure~\ref{fig:mlp} shows the MLP model we use for corneal data with three hidden layers. The network architecture is 29-128-256-4, with $29$ the dimension of the input data (i.e. the number of the variables of corneal data) and $128, 256$ and $4$ the numbers of neurons in the hidden layers in turn.

\subsection{Experimental results and clinical interpretation}
The MLP we utilise is a three-layer fully connected neural network which learns the diagnosis of A-K classification from the training data. The trained MLP model can be used to find the A-K classification label of the test data set. Of totally 237 samples, we use 72\% and 18\% in training and validation and 10\% in the test. We ran the experiments for 100 repetitions to reduce the randomness due to random shuffle of the data set and model training. We use 100 epochs for each training. Our MLP model demonstrated a mean validation accuracy of 73\% with epoch and variance of validation of 67-78\% across the analysis. The validation loss is decaying fast with a small variation. The test accuracy is 73.9\%. Figure 8 shows the ROC curves for the MLP model on each of four classes on corneal data with AUC values 0.92, 0.86, 0.83 and 0.91. The micro-average and macro-average AUC values (which reflect the general performance of the MLP classifier) are high at 0.87 and 0.90. The ROC curves are all close to the upper left corner, which indicates the excellent performance of the MLP classifier. We can thus infer that the MLP model demonstrates promising potential as an accurate artificial intelligence diagnosis for corneal diseases.

\section{Conclusion and discussion}\label{sec:conclusion_discussion}
We utilised an MLP model to learn the corneal disease diagnosis from labelled training data, and then developed a Bayesian neural network (Variational Autoencoder plus a Gaussian mixture model) to determine the degree of corneal disease from unlabelled corneal data. Both methods achieve a state of the art
performance on real corneal data.

In this study, the unsupervised VAE model resulted in higher accuracy than the MLP model. Training against an existing classification system, that is the A-K classification system in this study, is not needed with the unsupervised model. There are already multiple KC classification systems \cite{13-kovacs2016accuracy,14-yousefi2018keratoconus} utilised in research and clinical settings, resulting in inconsistency through research output and clinical interpretation. This VAE model with the 29-dimensional input could thus represent a potential independent and standardised classification system for KC.

As VAE results (see Figure~\ref{fig:vae_result_2}) showed, from left to right, 4 classes distinguished the patients from early to late stages, which fits well with the progression pattern of the disease. The cluster centre between classes 1 and 2 was closer to each other, suggesting less differentiation between features and variables of these groups. It is consistent with the clinical observation that early stages of KC are hard to distinguish. Furthermore, this clustering map seems to not only classify patients similarly to the A-K system, but it also visualises how much closer the features of patients (represented by the dots) are towards the next stage.  
Identifying where a patient may sit within their classification, that is at the beginning, middle or upper extremity may have significant additional impact contributing to the analysis of progression.

Accordingly, the next step in our research is to expand the model to include control samples inclusive of longitudinal outcomes.

Unlike research to date, a vital goal of this study has been not only to identify whether a patient has KC or not, but rather to be able to accurately group patients within the framework of the existing A-K KC classification stages.

Also of significance, the difference between this study and those reviewed is that we looked at both eyes of KC patients, as compared to the studies \cite{13-kovacs2016accuracy} and \cite{12-hwang2018distinguishing} which sought to identify sub-clinical KC in the fellow eyes of unilateral KC patients.

This study is to be able to accurately group patients within the framework of the existing A-K KC classification stages. We focus on clinician involvement in determining the variables considered and classification, which is an essential difference between our study and those of \cite{12-hwang2018distinguishing,13-kovacs2016accuracy,14-yousefi2018keratoconus}, which use built-in automated algorithms. 

While this study did not achieve the same sensitivity and specificity as the unsupervised model in \cite{14-yousefi2018keratoconus}, it is seeking to distinguish different stages within KC, whereas the high accuracy generated in \cite{14-yousefi2018keratoconus} was to differentiate between KC and control. Besides, our study represents a small data set compared to \cite{14-yousefi2018keratoconus}.

This study was able to achieve mean accuracy levels of 73\% and 80\% for supervised and unsupervised models respectively, and we expect that the inclusion of control samples alongside the existing clinical data may lead to additional improvement in these outcomes. It is important to continue training and testing the models to develop an approach which assists clinicians better manage KC and predict disease progression.

%\subsubsection*{Acknowledgments}

\ifCLASSOPTIONcaptionsoff
  \newpage
\fi

%\ifCLASSOPTIONcaptionsoff
%  \newpage
%\fi
%
%% trigger a \newpage just before the given reference
%% number - used to balance the columns on the last page
%% adjust value as needed - may need to be readjusted if
%% the document is modified later


%\bibliographystyle{IEEEtran}
\bibliographystyle{IEEEtran}
\bibliography{cornea}

\begin{thebibliography}{10}
\providecommand{\url}[1]{#1}
\csname url@samestyle\endcsname
\providecommand{\newblock}{\relax}
\providecommand{\bibinfo}[2]{#2}
\providecommand{\BIBentrySTDinterwordspacing}{\spaceskip=0pt\relax}
\providecommand{\BIBentryALTinterwordstretchfactor}{4}
\providecommand{\BIBentryALTinterwordspacing}{\spaceskip=\fontdimen2\font plus
\BIBentryALTinterwordstretchfactor\fontdimen3\font minus
  \fontdimen4\font\relax}
\providecommand{\BIBforeignlanguage}[2]{{%
\expandafter\ifx\csname l@#1\endcsname\relax
\typeout{** WARNING: IEEEtran.bst: No hyphenation pattern has been}%
\typeout{** loaded for the language `#1'. Using the pattern for}%
\typeout{** the default language instead.}%
\else
\language=\csname l@#1\endcsname
\fi
#2}}
\providecommand{\BIBdecl}{\relax}
\BIBdecl

\bibitem{1-date2019applications}
R.~C. Date, S.~J. Jesudasen, C.~Y. Weng \emph{et~al.}, ``Applications of deep
  learning and artificial intelligence in retina,'' \emph{International
  Ophthalmology Clinics}, vol.~59, no.~1, pp. 39--57, 2019.

\bibitem{2-motozawa2019optical}
N.~Motozawa, G.~An, S.~Takagi, S.~Kitahata, M.~Mandai, Y.~Hirami, H.~Yokota,
  M.~Akiba, A.~Tsujikawa, M.~Takahashi \emph{et~al.}, ``Optical coherence
  tomography-based deep-learning models for classifying normal and age-related
  macular degeneration and exudative and non-exudative age-related macular
  degeneration changes,'' \emph{Ophthalmology and Therapy}, pp. 1--13, 2019.

\bibitem{3-ahn2018deep}
J.~M. Ahn, S.~Kim, K.-S. Ahn, S.-H. Cho, K.~B. Lee, and U.~S. Kim, ``A deep
  learning model for the detection of both advanced and early glaucoma using
  fundus photography,'' \emph{PLOS One}, vol.~13, no.~11, p. e0207982, 2018.

\bibitem{4-serdarogullari2013prevalence}
H.~Serdarogullari, M.~Tetikoglu, H.~Karahan, F.~Altin, and M.~Elcioglu,
  ``Prevalence of keratoconus and subclinical keratoconus in subjects with
  astigmatism using {Pentacam} derived parameters,'' \emph{Journal of
  Ophthalmic \& Vision Research}, vol.~8, no.~3, p. 213, 2013.

\bibitem{5-tur2017review}
V.~M. Tur, C.~MacGregor, R.~Jayaswal, D.~O'Brart, and N.~Maycock, ``A review of
  keratoconus: diagnosis, pathophysiology, and genetics,'' \emph{Survey of
  Ophthalmology}, vol.~62, no.~6, pp. 770--783, 2017.

\bibitem{6-davidson2014pathogenesis}
A.~Davidson, S.~Hayes, A.~Hardcastle, and S.~Tuft, ``The pathogenesis of
  keratoconus,'' \emph{Eye}, vol.~28, no.~2, p. 189, 2014.

\bibitem{mcmonnies2015inflammation}
C.~W. McMonnies, ``Inflammation and keratoconus,'' \emph{Optometry and Vision
  Science}, vol.~92, no.~2, pp. e35--e41, 2015.

\bibitem{gordon2015risk}
A.~Gordon-Shaag, M.~Millodot, I.~Kaiserman, T.~Sela, G.~Barnett~Itzhaki,
  Y.~Zerbib, E.~Matityahu, S.~Shkedi, S.~Miroshnichenko, and E.~Shneor, ``Risk
  factors for keratoconus in {Israel}: a case--control study,''
  \emph{Ophthalmic and Physiological Optics}, vol.~35, no.~6, pp. 673--681,
  2015.

\bibitem{8-li2012anterior}
H.~Li, V.~Jhanji, S.~Dorairaj, A.~Liu, D.~S. Lam, and C.~K. Leung, ``Anterior
  segment optical coherence tomography and its clinical applications in
  glaucoma,'' \emph{Journal of Current Glaucoma Practice}, vol.~6, no.~2,
  p.~68, 2012.

\bibitem{7-martinez2017new}
A.~Mart{\'\i}nez-Abad and D.~P. Pi{\~n}ero, ``New perspectives on the detection
  and progression of keratoconus,'' \emph{Journal of Cataract \& Refractive
  Surgery}, vol.~43, no.~9, pp. 1213--1227, 2017.

\bibitem{ozer2019long}
M.~D. Ozer, M.~Batur, S.~Mesen, S.~Tekin, and E.~Seven, ``Long-term results of
  accelerated corneal cross-linking in adolescent patients with keratoconus,''
  \emph{Cornea}, vol.~38, no.~8, pp. 992--997, 2019.

\bibitem{15-li2009keratoconus}
X.~Li, H.~Yang, and Y.~S. Rabinowitz, ``Keratoconus: classification scheme
  based on videokeratography and clinical signs,'' \emph{Journal of Cataract \&
  Refractive Surgery}, vol.~35, no.~9, pp. 1597--1603, 2009.

\bibitem{10-alio2017keratoconus}
J.~L. Alio \emph{et~al.}, ``Keratoconus,'' \emph{Recent Advances in Diagnosis
  and Treatment}, 2017.

\bibitem{11-lin2019review}
S.~R. Lin, J.~G. Ladas, G.~G. Bahadur, S.~Al-Hashimi, and R.~Pineda, ``A review
  of machine learning techniques for keratoconus detection and refractive
  surgery screening,'' in \emph{Seminars in Ophthalmology}.\hskip 1em plus
  0.5em minus 0.4em\relax Taylor \& Francis, 2019, pp. 1--9.

\bibitem{12-hwang2018distinguishing}
E.~S. Hwang, C.~E. Perez-Straziota, S.~W. Kim, M.~R. Santhiago, and J.~B.
  Randleman, ``Distinguishing highly asymmetric keratoconus eyes using combined
  {Scheimpflug} and spectral-domain {OCT} analysis,'' \emph{Ophthalmology},
  vol. 125, no.~12, pp. 1862--1871, 2018.

\bibitem{13-kovacs2016accuracy}
I.~Kov{\'a}cs, K.~Mih{\'a}ltz, K.~Kr{\'a}nitz, {\'E}.~Juh{\'a}sz,
  {\'A}.~Tak{\'a}cs, L.~Dienes, R.~Gergely, and Z.~Z. Nagy, ``Accuracy of
  machine learning classifiers using bilateral data from a {Scheimpflug} camera
  for identifying eyes with preclinical signs of keratoconus,'' \emph{Journal
  of Cataract \& Refractive Surgery}, vol.~42, no.~2, pp. 275--283, 2016.

\bibitem{14-yousefi2018keratoconus}
S.~Yousefi, E.~Yousefi, H.~Takahashi, T.~Hayashi, H.~Tampo, S.~Inoda, Y.~Arai,
  and P.~Asbell, ``Keratoconus severity identification using unsupervised
  machine learning,'' \emph{PLOS One}, vol.~13, no.~11, p. e0205998, 2018.

\bibitem{kamiya2014evaluation}
K.~Kamiya, R.~Ishii, K.~Shimizu, and A.~Igarashi, ``Evaluation of corneal
  elevation, pachymetry and keratometry in keratoconic eyes with respect to the
  stage of {Amsler-Krumeich} classification,'' \emph{British Journal of
  Ophthalmology}, vol.~98, no.~4, pp. 459--463, 2014.

\bibitem{16-kingma2013autoencoding}
D.~P. Kingma and M.~Welling, ``Auto-encoding variational {Bayes},'' in
  \emph{ICLR}, 2014.

\bibitem{17-dilokthanakul2016deep}
N.~Dilokthanakul, P.~A.~M. Mediano, M.~Garnelo, M.~C.~H. Lee, H.~Salimbeni,
  K.~Arulkumaran, and M.~Shanahan, ``Deep unsupervised clustering with
  {Gaussian} mixture variational autoencoders,'' 2016.

\bibitem{mackay2003information}
D.~J. MacKay, \emph{Information theory, inference and learning
  algorithms}.\hskip 1em plus 0.5em minus 0.4em\relax Cambridge University
  Press, 2003.

\bibitem{kullback1951information}
S.~Kullback and R.~A. Leibler, ``On information and sufficiency,'' \emph{The
  Annals of Mathematical Statistics}, vol.~22, no.~1, pp. 79--86, 1951.

\bibitem{18-lecun2015deep}
Y.~LeCun, Y.~Bengio, and G.~Hinton, ``Deep learning,'' \emph{Nature}, vol. 521,
  no. 7553, pp. 436--444, 2015.

\bibitem{19-Goodfellow-et-al-2016}
I.~Goodfellow, Y.~Bengio, and A.~Courville, \emph{Deep Learning}.\hskip 1em
  plus 0.5em minus 0.4em\relax MIT Press, 2016,
  \url{http://www.deeplearningbook.org}.

\end{thebibliography}

%%%%%%%%%%%%%%%%%%%%%%%%%%%%%%%%%%%%%%%%%%%%%%%%%%%%%%%%%%%%%%%%%%%%%%%%%%%%%%%%%%%%%%%%%%

%\newpage

\end{document}